\RequirePackage{fix-cm}
\documentclass[smallextended]{svjour3}       
\smartqed  
\usepackage{color}
\usepackage{graphicx}
\usepackage{amsmath}
\usepackage{booktabs}
\usepackage{multirow}
\newcommand{\rev}[1]{\textcolor{black}{#1}}
\newcommand\dunderline[3][-2pt]{{%
  \setbox0=\hbox{#3}
  \ooalign{\copy0\cr\rule[\dimexpr#1-#2\relax]{\wd0}{#2}}}}
\begin{document}

\title{Integrating Combined Task and Motion Planning with Compliant Control}
\subtitle{Successfully conducting planned dual-arm assembly motion using compliant peg-in-hole control}

\author{Hao Chen \and Juncheng Li \and Weiwei Wan* \and Zhifeng Huang \and Kensuke Harada 
}


\institute{Hao Chen, Weiwei Wan, and Kensuke Harada  \at
              Graduate School of Engineering Science, Osaka University, Japan\\
              Correspondent author: Weiwei Wan, Tel.: +810368506381, x\email{wan@sys.es.osaka-u.ac.jp}           
           \and
           Juncheng Li and Zhifeng Huang\at
              School of Automation, Guangdong University of Technology, China
}

\date{Received: date / Accepted: date}

\maketitle

\begin{abstract}
Planning a motion for inserting pegs remains an open problem. The difficulty lies in both the inevitable errors in the grasps of a robotic hand and absolute precision problems in robot joint motors. This paper proposes an integral method to solve the problem. The method uses combined task and motion planning to \rev{plan} the grasps and motion for a dual-arm robot to pick up the objects and move them to assembly poses. Then, it controls the dual-arm robot using a compliant strategy (a combination of linear search, spiral search, and impedance control) to finish up the insertion. The method is implemented on a dual-arm Universal Robots 3 robot. \rev{Six objects, including a connector with fifteen peg-in-hole pairs for detailed analysis and other five objects with different contours of pegs and holes for additional validation, were tested by the robot.} Experimental results show reasonable force-torque signal changes and end-effector position changes. The proposed method exhibits high robustness and high fidelity in successfully conducting planned peg-in-hole tasks. 
\end{abstract}

\section{Introduction}
\label{intro}
Assembly, especially peg-in-hole is the essential problem as well as the holy grail in robotic systems for industrial automation. Although peg-in-hole has been studied for many years, the complexity in object shapes and connecting mechanism make the peg-in-hole the problem remains the interest of many robotic researchers. \rev{Conventional} peg-in-hole requires well-prepared fixture settings and limited initial configurations, making the deployment of peg-in-hole robotic systems in mass production difficult. Most of them have an assumption that the peg is firmly held by a robot hand and the hole is fixed in the environment. On the other hand, recent robotic systems tend to use high-level interface like Robotic Operation System (ROS) \cite{quigley2009ros} to generate robot motion. Visual recognition is used to detect the initial pose of objects. Probabilistic roadmap methods \cite{Thierry04} are used to generate the robot motion. The automatic recognition and motion planning significantly reduces the cost needed in the deployment of robotic systems in industrial automation. Previously, the recognition and planning problem was considered to be \rev{kinematic}. The start is one or multiple joint configurations that reaches the initial object pose. The goal is one or multiple joint configurations at the goal object pose. The motion planning finds a feasible joint sequence between the start and goal. Most of the previous research on recognition and planning were carried out for pick-and-place tasks \cite{cowley2013perception}\cite{domae2014fast}\cite{harada2014validating}. Planning a motion for inserting pegs is less studied and remains an open problem. The difficulty lies in the inevitable errors in the grasps of a robotic hand and absolute precision problems in robot joint motors. These errors and problems may lead to the failure in the peg-in-hole assembly that follows the planned motion.

\begin{figure}[!htpb]
  \begin{center}
  \includegraphics[width=.75\linewidth]{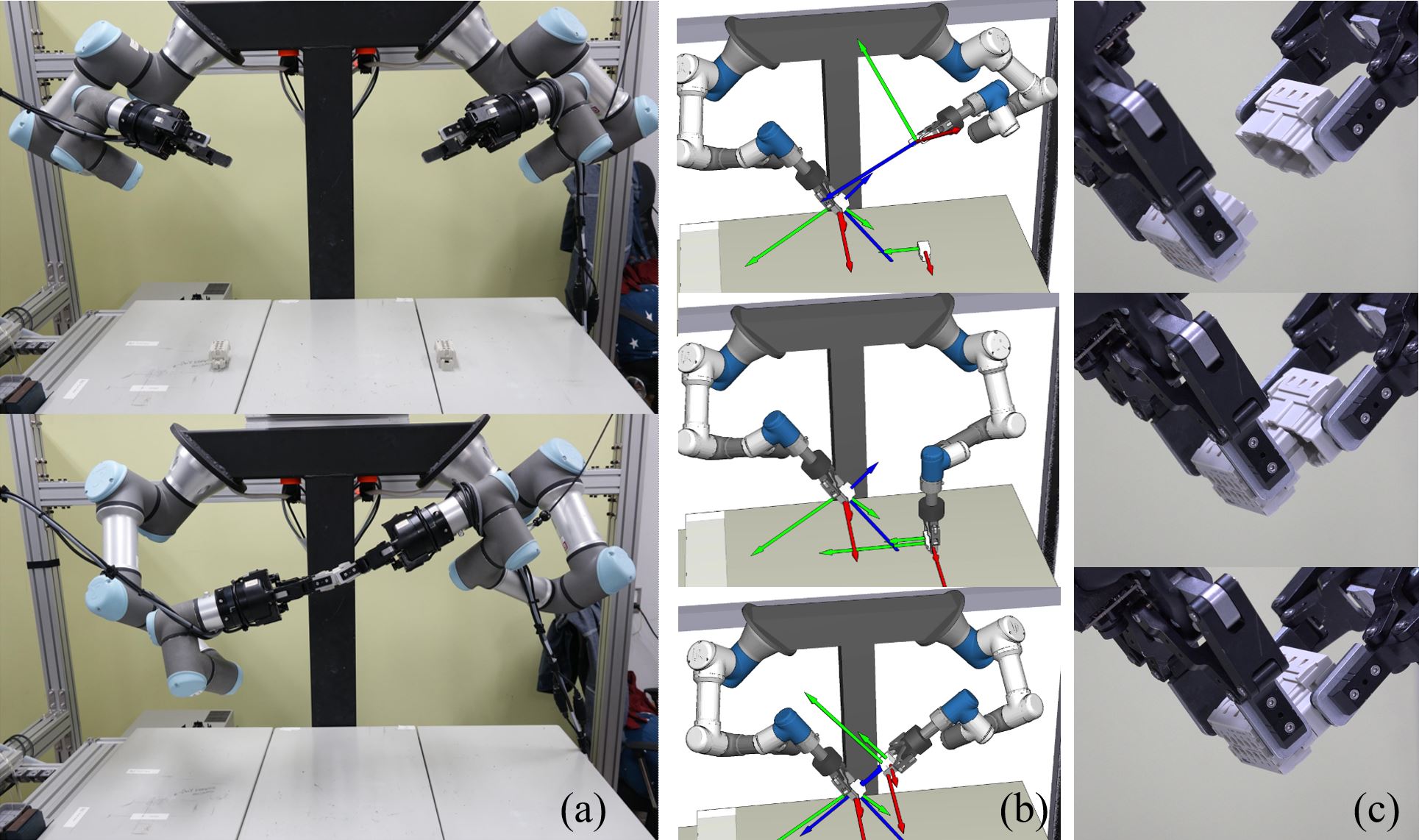}
  \caption{(a) At the starting point, the objects are randomly placed on a table. (b) The system automatically plans the grasps and motion for a dual-arm robot to pick up the objects and move them to assembly poses. (c) The system controls the dual-arm robot using a compliant strategy to finish up the intertion.}
  \label{fig:teaser}
  \end{center}
\end{figure}

In this paper, we proposed a method to successfully conduct planned dual-arm assembly motion using compliant peg-in-hole control. The goal of our study is to use a dual-arm robot to perform the fully automatic peg-in-hole assembly. Fig.\ref{fig:teaser} illustrates our problem setting. At the starting point, the objects are randomly placed on a table. There are no special requirements for fixtures or initial object poses. Our system automatically plans the grasps and motion for a dual-arm robot to pick up the objects and move them to assembly poses. Then, it controls the dual-arm robot using a compliant strategy to finish up the insertion. In the end, the objects are expected to be assembled and held by one of the robot hands for later usage. 

\rev{Our main contrition is in the integration of combined task and motion planning and compliant control. We integrally solve the manipulation planning of the objects and the force control of the insertion. In conventional peg-in-hole tasks, the position of a hole was mostly assumed to be fixed and a robot conducted the assembly task relative to the hole. In contrast, our study enables a robot to flexibly choose the poses of the assembly counterparts and thus increases the feasibility of the assembly task.}

The proposed method is implemented on a dual-arm Universal Robots 3 robot. \rev{Six objects, including a connector (ten circular peg-hole pairs, three rectangular peg-hole pairs, and two trapezoid peg-hole pairs) with fifteen peg-in-hole pairs for detailed analysis and other five objects (a D-sub DB25 connector, a usb connector, a RJ45 internet connector, a one-touch vacuum tube connector, and a three-prong power plug) with different contours of pegs and holes for additional validation, were tested by the robot.} Experimental results show reasonable force-torque signal changes and end-effector position changes. The proposed method exhibits high robustness and high fidelity in successfully conducting planned peg-in-hole tasks. 

The organization of the paper is as follows. Section II reviews the related studies. Section III shows an overview of the developed system and method. Section IV and Section V respectively discuss in detail the planning algorithms and peg-in-hole control strategies. Section VI presents experiments and analysis. Conclusions are drawn in Section VII.

\section{Related work}

The related work includes peg-in-hole assembly, combined task and motion planning, and dual-arm robotic assembly.

\subsection{Peg-in-hole}
A peg-in-hole problem consists of two phases: search and insertion. The search phase is to find the hole in the assembly object. The insertion phase is to insert the mating object into the hole in the assembly object. Different researches have already done in these two different phases.

\subsubsection{Search phase}
The search phase runs before the actual insertion. The goal of this phase is to deal with large offsets between the peg and the hole. 

The early search was done with the help of visual surveillance. \rev{They required the feedback information from vision sensors and control the motion of the robot based on it.} For example, Shirai \rev{et al.}\cite{shirai1973guiding} used the visual feedback to insert a square peg into a square hole. The color of the bottom of the square hole is different and the camera finds the position by detecting the bottom. The work required very clear boundary information and the shape of the peg was limited to simple geometries. \rev{Yoshimi et al. \cite{yoshimi1994active} proposed an uncalibrated visual servoing method which reasons control information directly from an image. Huang et al. \cite{huang2013fast} used high-speed cameras to implement fast searching and aligning. Song et al. \cite{song2014automated} considered the contour of the peg models when aligning the peg to the hole. More contemporary methods use force feedback and compliance-based strategies \cite{li2017human} for searching. For example, Newman et al. \cite{newman2001interpretation} proposed the force guide map together with random or spiral tries to locate holes. Chhatpar et al. \cite{chhatpar2001search} proposed a blind search strategy that can search the hole within a certain circular area of a specific radius. Park et al. \cite{park2017compliance} proposed a hole search strategy that conducted the human behaviour motion like tilting, through the analysis of several contact states. Marvel et al. \cite{marvel2018multi} studied the spiral search and its variants under multi-robot coordination. Abdullah et al. \cite{abdullah2015approach} proposed a strategy to find the location of the hole by evaluating the reaction moments at the contact.}

The search in our work is a combination of spiral tries and tilting. They work together with compliance and impedance-based insertion to conduct successful assembly.

\subsubsection{Insertion phase} 
In the insertion phase, a peg is considered to be partially inside a hole, and the goal is to successfully finish up the insertion. Researchers used the compliance-based method to solve the problem. The compliance-based methods are divided into two categories: One uses mechanical passive compliance; The other practices active compliance with the help of various feedback. The passive category usually uses a mechanism named remote center compliance(RCC) \cite{whitney1982quasi} to provide compliance to the peg. The mechanism could be installed on either the robot wrist or palm.
It makes the relation of peg and gripper fixed and meanwhile provides the flexibility to the peg during the insertion. The very first RCC was proposed by Whitney et al. in \cite{whitney1982quasi}. The mechanism design can do the peg-hole insertion task without jamming. Many similar mechanisms were proposed after that. For example, the work done by Haskiya \cite{haskiya1998passive} built a vertical/horizontal remote center compliance mechanism that can work in both vertical and horizontal directions. Also, a different method is proposed to overcome the fixed position of the compliant center and fixed stiffness to expand assembly tasks flexibility without additional configuration cost
\cite{zhao1998vrcc}\cite{joo1998development}\cite{lee2005development}. More recently, Suzuki el.at \cite{dreaming} designed an RCC with a push-activate-rotation function. It can rotate around a vertical axis by pushing the wrist in the vertical direction. The active category designs a contact model for peg-hole insertion and finds an insertion strategy that can minimize the contact error. For example, Balletti et al. \cite{balletti2012towards} and Song et al. \cite{song2016guidance} used force sensing feedback and impedance control. The impedance control theory proposed by Hogan \cite{hogan1987stable} is widely used in such studies. Impedance control builds a connection between the force and motion of the peg so that the robot can align the peg by flexible adjustment. Some other methods use combined force-vision feedback. Liu et al. \cite{liu2019} combined the force vision system and force information to supervise the visually measurable compliance to improve the alignment process. Similar methods were used by Nammoto et al. \cite{nammoto2013model} to assemble 2D objects. More recently, researchers tend to use machine learning-based methods, which is expected to exhibit adaptation in the presence of varying object shapes.

The peg-in-hole method used in this paper is the active compliance one. Compliance-based assembly strategies together with impedance control are used to deal with complex connector shapes.

\subsection{Combined task and motion planning}

Combined task and motion planning is used to simultaneously generate action sequence and motion sequence for a given task. It has two levels of planning which interact with each other. In the high level, the planning is symbolic deduction \cite{wally2019flexible}\cite{zhang2017multirobot}. It is essentially a series of logic operations to determine a sequence of action-level strategies. In the low level, the planning is topological optimization \cite{ratliff2009chomp}\cite{schulman2014motion} or graph search \cite{Thierry04}\cite{Lavalle00}.
The two levels switch back and forth in the presence of failures.

One seminal study of combine task and motion planning was presented by Wolfe et al. \cite{wolfe2010combined}. The idea is inherited and extended by several following studies like \cite{Kaelbling13}\cite{garrett2015ffrob}\cite{woosley2018integrated}. Our group at Osaka University also started developing combined task and motion planners since 2015 \cite{Wan2015a}. The planner can be used by dual-arm robots to automatically determine pick-and-place and handover \cite{wan2019tii}. It can also be used to plan assembly motion using single-arm regrasp\cite{Wan2016tdar} or dual-arm collaboration\cite{moriyama2019dual}. This paper is built on the developed planner. It extends the dual-arm collaboration in \cite{moriyama2019dual} with a compliant peg-in-hole strategy to deal with realistic tasks.

\subsection{Dual arm robot assembly}
The dual-arm robot assembly is receiving attention in the recent two years. Previous, there were lots of studies relating to dual-arm manipulation \cite{siciliano2012advanced}\cite{kruger2011dual}. Most of them were concentrated on planning sequential \cite{cohen2015planning}\cite{kurosu2017simultaneous} or coordinated motion \cite{ramirez2017human}. Practical tasks in industrial automation using dual-arm robots are less studied \cite{hwang2007motion}.  Recently, with the decrease in cost, using two robots, instead of a robot and a fixture to perform assembly tasks is getting into focus. Some examples include \cite{stavridis2018bimanual}\cite{moriyama2019dual}. One of the robotic arms in these studies is used as a general fixture. It holds one of the target objects and waits for the mating action of the other arm.

Different from the aforementioned dual-arm robotic assembly studies, we perform dual-arm assembly considering the ensuing insertion problems. Insertion constraints are integrally analyzed and incorporated. \rev{The dual-arm robot is more advantageous in assembly and insertion since it can control the relative motion and interaction of assembly counterparts in a dexterous human-like manner. Dual arms allow the robot to conduct handover and reorientate the object, thus increasing the dexterity of manipulation and feasibility of planning. Although a dual-arm robot is difficult to control and has relatively high costs, with the help of the algorithms and methods proposed in this paper, it could obtain significant dexterity and flexibility, exhibiting well-deserved cost-performance.}

\section{System overview}

\begin{figure}[!htpb]
  \begin{center}
  \includegraphics[width=.95\linewidth]{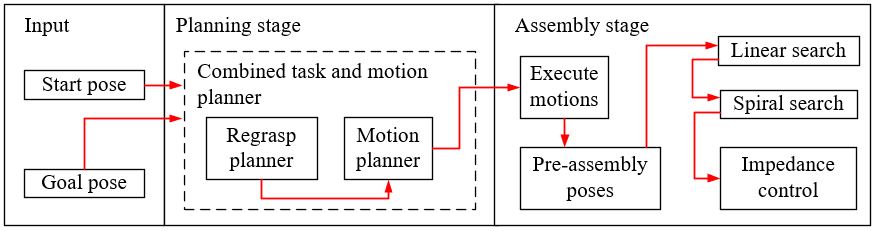}
  \caption{\rev{An overview of the system.}}
  \label{fig:flowchart}
  \end{center}
\end{figure}

This section an overview of the developed system and method. Fig.\ref{fig:flowchart} illustrates the workflow.

The two objects used for assembly are named the assembly object and the mating object respectively. In the beginning, a vision system detects the initial poses of the assembly object and the mating object. Meanwhile, human workers specify the final assembly poses of the two objects. The initial pose and goal pose will be used by a combined task and motion planner in the planning state to generate a sequence of actions. The combined planner automatically computes the IK-feasible and collision-free path for the robot to move the assembly art and the mating object from their initial poses to the pre-assembly poses, which are defined as one-step before the finally assembled states. The robot moves the components to the pre-assembly poses to prepare for the compliant control.

In detail, the combined task and motion planner consists of two sub-planners -- a regrasp planner and a motion planner. The regrasp planner can generate a sequence of the grasps to orientate an object from its initial pose to goal pose. The motion planner can obtain a smooth and collision-free path between every two grasps.

After moving the objects to the pre-assembly poses, the system switches to compliance control mode (the assembly strategy section in Fig.\ref{fig:flowchart}). There are three compliant strategies -- Linear search, spiral search, and impedance control. The linear search is a linear movement for the robot. During a linear search, the robot arm moves towards a direction in a straight line until the reaction force is larger than a given threshold. The spiral search is a common search strategy for finding the hole in the assembly object. The robot arm that holds the mating object tries with spiral touching points to detect holes. After finding the hole, the robot arm will switch to impedance control. The details will be explained in the following two sections.

\section{Planning}

The planning strategy used to generate the motions for a robot to grasp the objects and move them to pre-assembly poses is based on the combined task and motion planning framework proposed by Wan et al. \cite{wan2019tii}. \rev{A diagram showing the planning workflow is presented in Fig.~\ref{fig:planningflowchart}. It} comprises a grasp planner at a high level and a motion planner at a low level. By using the initial pose and goal pose of the objects, the regrasp planner can generate a sequence of grasps for a robot to pick up the objects, orientate them, and move them to the goal pose. The motion planner can generate the smooth and collision-free motions to move between adjacent two grasps in the regrasp sequence.

\begin{figure}[!htpb]
  \begin{center}
  \includegraphics[width=.95\linewidth]{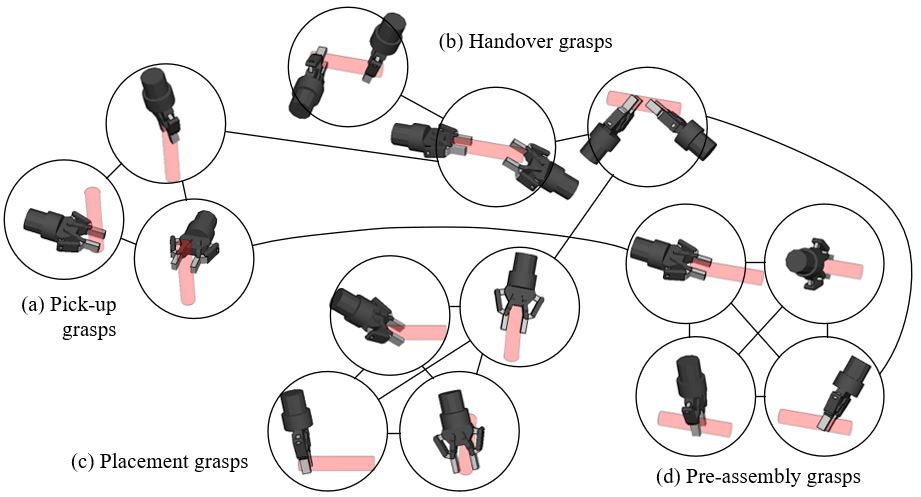}
  \caption{An illustration of the regrasp graph. Nodes in the group (a) indicate the plausible grasps for picking up an object from its initial pose. Nodes in (b) indicate the plausible grasps for handover.  Nodes in (c) indicate the plausible grasps to place down and pick up an object from placement on a flat surface. Nodes in the group (d) indicate the plausible goal grasps at the pre-assembly pose. The high-level component of the combined motion and task planner finds a sequence of grasps by building and searching the regrasp graph.}
  \label{fig:regrasp}
  \end{center}
\end{figure}

The regrasp planner determines a sequence of poses by generating a regrasp graph and searching the regrasp graph to find the shortest path in the graph. Fig.\ref{fig:regrasp} is an example of the regrasp graph. 

\rev{
\begin{figure}[!htpb]
  \begin{center}
  \includegraphics[width=.95\linewidth]{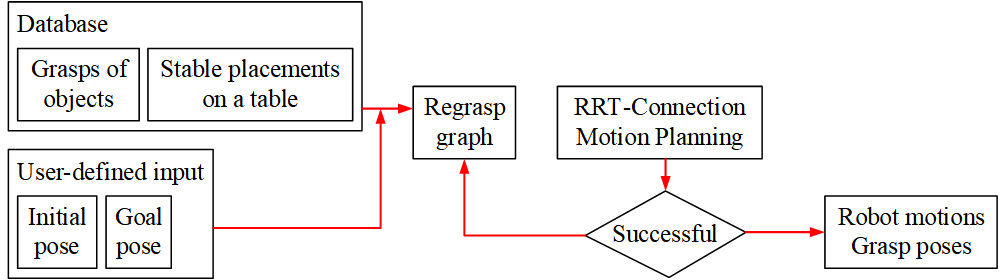}
  \caption{The planning workflow.}
  \label{fig:planningflowchart}
  \end{center}
\end{figure}
}

The nodes in the group (a) of the graph represent all the collision-free and IK-feasible grasps for the object at an initial pose. The nodes in the group (d) of the graph represent all the collision-free and IK-feasible grasps for the object at the goal. The nodes in the group (b) are the plausible handover grasps for the object. The nodes in the group (c) are the plausible grasps for stationary poses (placements) on flat surfaces. After generating the regrasp graph, the regrasp planner searches the regrasp graph and tries to find the shortest path in the graph. In this process, the planner firstly verifies if there exists a grasp that can both pick up an object at its initial pose (nodes in the left of the graph) as well as hold the object at its goal pose (nodes in the right of the graph). The existence of such an object means the robot can directly move the object from the initial pose to the goal pose. It does not need handover or placement motion to reorientate the object. If the grasp does not exist, the planner finds some handover or placement grasps from the group (b) and (c), and use them to generate a sequence of handover grasps or pick-up and place-down to move the object from the initial pose to the goal pose. Interested readers are encouraged to read the previous work \cite{wan2019tii} for details.

The motion planner generates the motions between every two grasps in the path found in the regrasp graph. Probabilistic motion planning methods like RRT-connection are used to generate the motion.  \rev{RRT-connection} traverses a configuration space by generating two random search trees from both the start and the goal. The two trees randomly sample a direction in the configuration space and append a node towards the random direction with the shortest distance to the tree. The newly appended point will be the end node of the tree. The path is found when the distance between the end nodes of the two trees is smaller than a threshold. The algorithm unnecessarily finds a path. When there is no path found, the edge between the two grasps in the regrasp graph is deleted. The motion component backtracks to the regrasp component and re-searches the regrasp graph to find a new grasp sequence and plan the motion again. If all edges are deleted and no results are planned, the planner will remove both the node and the edge of the grasps in the old pose and rebuild the edge and the node in \rev{a new pose}. The object is treated as a stationary obstacle in motion planning.

To reduce the uncertainty in the peg-in-hole assembly process, 
we add special constraints to the handover process to reduce the uncertainty of the mating object. The constraints force the planner to include at least one perpendicular handover so that one grasp corrects the pose error in the one local axis and another grasp corrects the pose error in another local axis. The two perpendicular handover grasps can eliminate the error of the mating object. For the assembly object, since one hand is already holding an object, handover is no longer applicable. Compliant control is further employed.

\section{Compliant Strategy}

As mentioned previously, for the mating object, the robot uses handover to eliminate the error in the rotation of the mating object. After that, the robot grasps the assembly object and moves it to its assembly pose. Since the first hand is already grasping the mating object\rev{, t}he left-arm may only change the pose of the second object using pick-and-place based regrasp. The motion could not eliminate the rotation error of the assembly object. \rev{For these reasons, we assume the errors include both the positional and rotational ones (could be caused by calibration or uncertainty in pick and place).}

\begin{figure}[!htpb]
  \begin{center}
  \includegraphics[width=.95\linewidth]{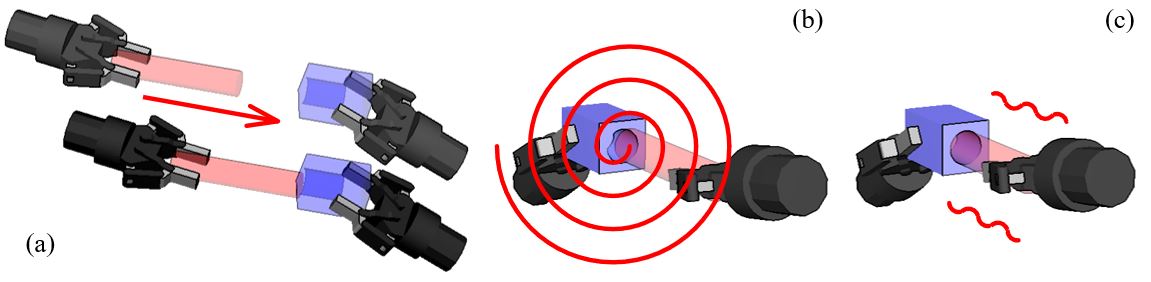}
  \caption{Compliant strategies used to finish up the insertion. (a) Linear search.
  (b) Spiral search (c) Impedance control.}
  \label{fig:control}
  \end{center}
\end{figure}

Consequently, after moving two objects to their pre-assembly poses, we fix the direction of the mating object, and divide the compliant assembly strategy into three stages. In the first stage, the mating object is controlled to move toward the assembly object. It will stop until the force sensor detects the force change (the mating object is in touch with the assembly object). Then, the robot performs a spiral search to locate a hole. In the third stage, the robot employs impedance control to overcome the uncertainty in the rotation. The three stages are illustrated in Fig.\ref{fig:control}. Please see the subsections for details.

\subsection{Linear search and Spiral search}

The goal of the linear and spiral search is to find the object and locate the hole. Both of them require an F/T sensor to be installed at the wrist.

During the linear search, a robot arm \rev{holding} a mating object moves along a given vector $\mathbf{v}^{direction}$ until the following equation is \rev{violated}, as is shown in Fig.\ref{fig:control}(a).
\begin{equation}
  \mathbf{v}^{direction}\cdot(\mathbf{R}^{hnd}\cdot\mathbf{F})\leq threshold
  \label{linearstop}
\end{equation}
Here, $\mathbf{R}^{hnd}$ is the rotation of the holding hand. $\mathbf{F}$ is the force obtained using
an F/T sensor. $\mathbf{v}^{direction}$ is the insertion direction determined by the assembly pose.
$threshhold$ is defined as a value larger than the sensor noise.

Assume the robot hand stops at a position $\mathbf{P}^{hnd}_0$. During the spiral search, the robot arm maintains its $\mathbf{R}^{hnd}$, but changes $\mathbf{P}^{hnd}$ along spiral curves on a plane \rev{where the normal is the same as $\mathbf{v}^{direction}$}
. Here, the hand coordinate is defined following the conventional robotics textbook. 
The $i+1$th spiral position is computed using
\begin{equation}
  \mathbf{P}^{hnd}_{i+1}=r_{i+1}\cdot\mathtt{rodrigues}(\theta_{i+1}, \rev{\mathbf{v}^{direction}})\cdot\rev{(\mathbf{I}_{x}+\mathbf{I}_{y})}+\mathbf{P}^{hnd}_{i}
\end{equation}
where
\begin{equation}
\theta_{i+1} = \theta_{i}+\delta\theta,~r_{i+1} = r_{i}+\delta r
\end{equation}
$\delta\theta$ and $\delta r$ are the discretized step length of the spiral search.
\rev{$\mathbf{I}_{x}$ and $\mathbf{I}_{x}$ are the first and second column of an identity matrix $\mathbf{I}$, respectively.}
$\mathtt{rodrigues}(\theta, \mathbf{v})$ is the Rodrigues' rotation formula \rev{\cite{rodrigues1840lois}}. It is computed using
\begin{equation}
  \mathtt{rodrigues}(\theta, \mathbf{v}) = \mathbf{I}+sin\theta[\hat{\mathbf{v}}\times]+(1-cos\theta)[\hat{\mathbf{v}}\times]^2
\end{equation}
The spiral search stops until equation \eqref{linearstop} is violated.

\subsection{Impedance control}

Our impedance control is implemented in the workspace instead of the robot joint space. It follows the conventional impedance control law
\begin{equation}
  \mathbf{F}_i = m\cdot\mathbf{\ddot{P}}^{hnd}_i+c\cdot\mathbf{\dot{P}}^{hnd}_i+k\cdot(\mathbf{\rev{P}}^{hnd}_i-\mathbf{\rev{P}}^{hnd}_{\rev{i-1}})
\end{equation}
where $\mathbf{\ddot{P}}^{hnd}_i$, $\mathbf{\dot{P}}^{hnd}_i$, and $\mathbf{{P}}^{hnd}_i$ are the acceleration, speed, and position of the holding hand. $m$, $c$, and $k$ are the \rev{inertia}, damping coefficient, and \rev{position gain} respectively.

The $i+1$th position is thus updated by
\begin{equation}
  \mathbf{P}^{hnd}_{i+1} = \frac{\mathbf{F}_i+m\frac{(2\mathbf{P}^{hnd}_{i}-\mathbf{P}^{hnd}_{i-1})}{dt^2}+c\frac{\mathbf{P}^{hnd}_{\rev{i}}}{dt}+k\mathbf{P}^{hnd}_{\rev{i}}}{\frac{m}{dt^2}+\frac{c}{dt}+k}
\end{equation}
\rev{where $dt$ is the sampling time.} It will be used by the robot control in the insertion to adjust the mating object. 

\section{Experiments and Analysis}

\begin{figure}[!htpb]
  \begin{center}
  \includegraphics[width=.97\linewidth]{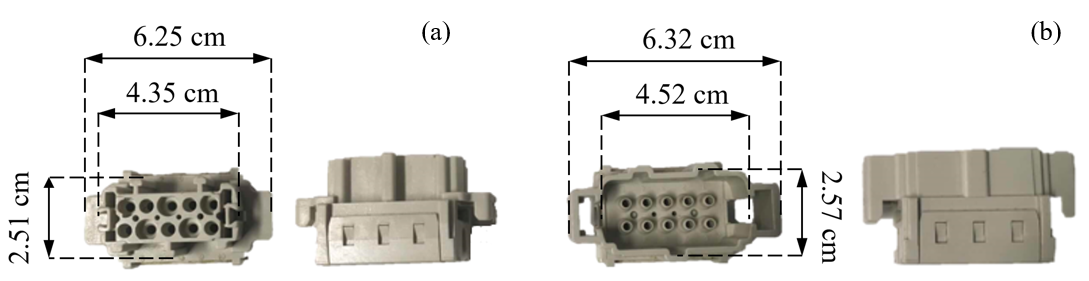}
  \caption{The mating object (a) and assembly object (b) used in the experiment. \rev{The left part of each subfigure is the top view of the object. The right part is the front view.}}
  \label{fig:twoparts}
  \end{center}
\end{figure}

The proposed method is validated using a dual-arm robot made of two Universal Robots 3 robotic arms. Each UR3 arm is equipped with Robotiq 85 gripper and Robotiq F/T300 force sensor. The configuration of the computer we used to do the planning has an Intel Core i9-9900K CPU,  four 16GB 3600MHz DDR4 memory and a GeForce GTX 1080Ti GPU. The assembly object and mating object used in the experiment are from a manifold connector shown in Fig.\ref{fig:twoparts}. The connect has ten circular peg-hole pairs in the center, three rectangular peg-hole pairs at the long edges, and two trapezoid peg-hole pairs at the short edges.

\begin{figure}[!htpb]
  \begin{center}
  \includegraphics[width=.9\linewidth]{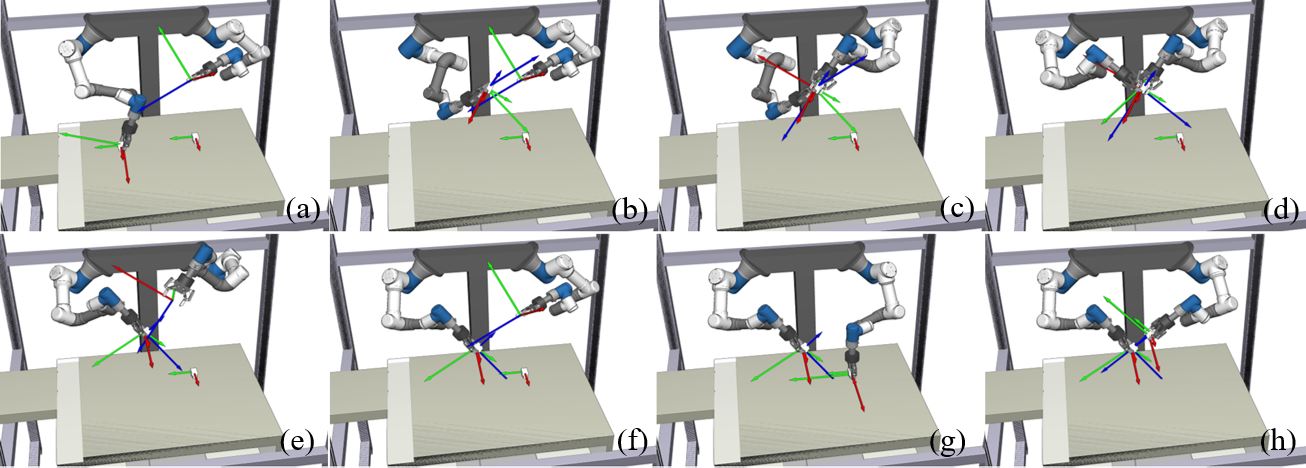}
  \caption{\rev{The planned motion in the simulation. The subfigures (a)-(h) show the motion sequence found by the planner to manipulate and assemble the objects.}}
  \label{fig:simulation}
  \end{center}
\end{figure}

A sequence of snapshots showing the planned motion in the simulation environment is presented in Fig.\ref{fig:simulation}. The robot chooses a grasp to pick the mating object up using its right arm in (a). Then, it performs a perpendicular handover in (b)-(e) to reduce the uncertainty. The object is handed over to the left arm in (c), oriented by the left arm in (d), and handed over back to the right arm in (e). In (g), the robot chooses a grasp for its left arm to pick up the assembly object. At the end of the simulation, both objects are moved to pre-assembly poses, as is shown in (h).

\begin{figure}[!htpb]
  \begin{center}
  \includegraphics[width=.8\linewidth]{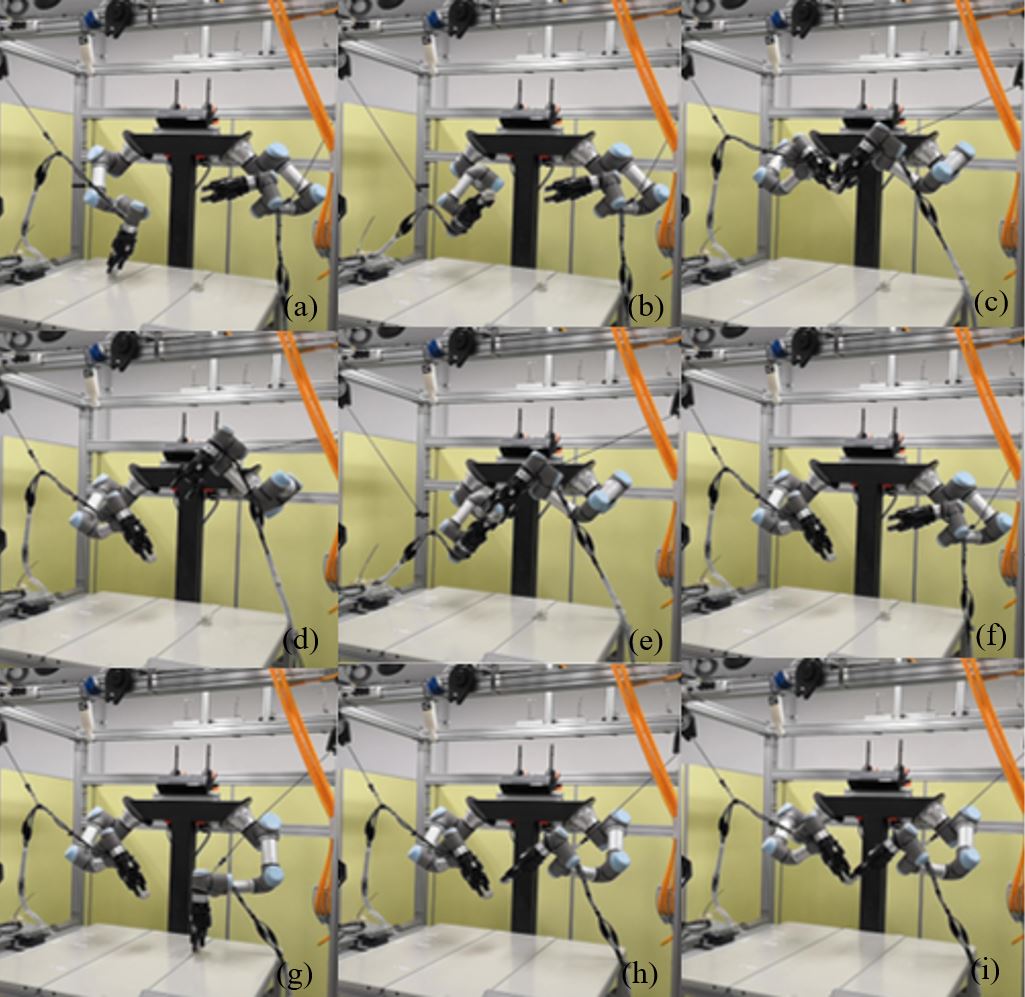}
  \caption{The planned motion in real execution. The subfigures (a)-(h) corresponds exactly to the motion planned in the simulation. Subfigure (i) shows the final state after applying compliant control.}
  \label{fig:execution}
  \end{center}
\end{figure}

Fig.\ref{fig:execution} shows the execution results of the planned motion. The two arms follow the planned sequence and motion to pick up and move the two objects. The subfigures (a)-(h) correspond exactly to the motion planned in the simulation. Subfigure (i) shows a finished state. The aforementioned assembly strategy is performed to finally assemble the objects from the states in (h) to (i). A video including simulation, execution, and the close-up control motion could be found in the supplementary material submitted together with the paper.
We encourage our readers to refer to the video for a better view of the results.

\begin{figure}[!htpb]
  \begin{center}
  \includegraphics[width=.99\linewidth]{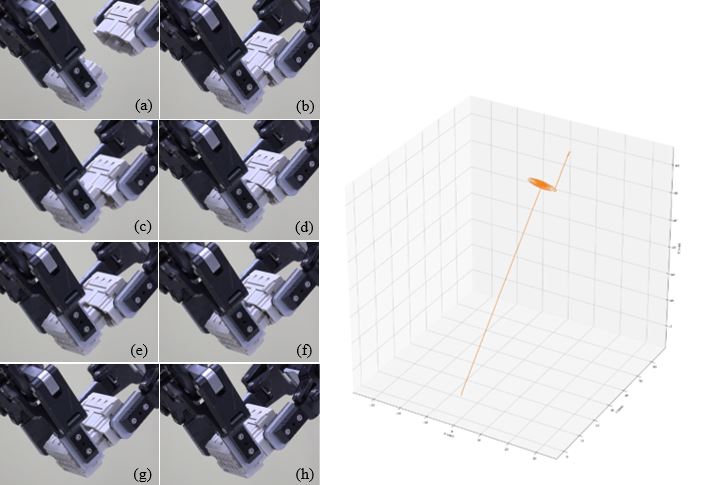}
  \caption{The changes in the two objects during the compliant peg-in-hole control process. The left pictures shows a view of the realworld. The right diagram shows the trajectory of the mating gripper.}
  \label{fig:complianceexe}
  \end{center}
\end{figure}

\begin{figure}[!htpb]
  \begin{center}
  \includegraphics[width=.99\linewidth]{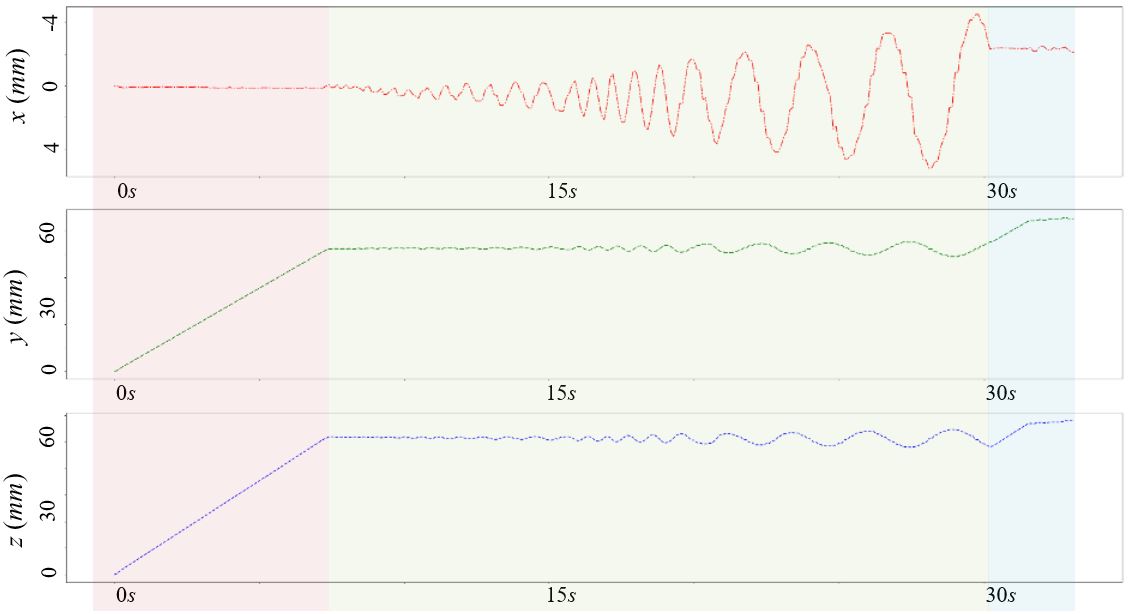}
  \caption{The changes in the $x$, $y$, $z$ positions of the mating gripper. The whole compliant control took around 30 $seconds$. The robot is performing linear search during the red time section, switches to spiral search in the green time section, and changes to impedances control in the blue time section.}
  \label{fig:xyz}
  \end{center}
\end{figure}

Fig.\ref{fig:complianceexe} shows the changes in the two objects during compliant peg-in-hole control. The robot is conducting a linear search in (a)-(b). Then, it switches to a spiral search in (b)-(f). The robot finds the hole at (g) and switches to impedance control mode. Finally, the mating object is inserted into the assembly object in (h). The trajectory of the mating gripper in the process is shown in the right object of Fig.\ref{fig:complianceexe}. The long segment indicates a linear search. The circles at the end of the long segments indicate a spiral search. The short segment indicates the impedance motion. The changes in the $x$, $y$, $z$ positions of the mating gripper are shown in Fig.\ref{fig:xyz}. The whole compliant control took around 30 $seconds$. \rev{The detailed time costs of the whole process are shown in Table.\ref{tb:timeconsuming}} The robot is performing linear search during the red time section and the $x$, $y$, $z$ positions are changing along straight lines. The robot switches to spiral search during the green time section. The $x$, $y$, $z$ positions, especially the $x$ positions, are moving back and forth along a wave.
The robot switches to impedance control in the blue time section.
The $x$, $y$, $z$ positions in this period \rev{change} following the impedance gains.

\begin{table}[!htbp]
  \centering
  \caption{\label{conf0}Time costs of each stage of the experiments}
  \label{tb:timeconsuming}
  \resizebox{0.75\linewidth}{!}{%
  \begin{tabular}{cc|c|c|c}\toprule
  \dunderline{.57pt}{Mating object} &  Pick up & Regrasp 1 & Regrasp 2  & Goal pose \\
  ~ & 7$s$         & 23$s$            & 22$s$               & 4$s$       \\ \midrule
  \dunderline{.57pt}{Assembly object} & Grasp & Goal pose & - &-  \\ & 7$s$ & 6$s$   & ~ &~  \\ \midrule
  \dunderline{.57pt}{Compiant control} & Linear & Spiral & Impedance  & - \\
  ~ & 8$s$        & 25$s$           & 3$s$ & ~  \\
  \bottomrule
  \end{tabular}}
\end{table}

An important issue of compliant peg-in-hole assembly is local minima. From a physical view, the robot gets stuck when it encounters a local minimum. For the given objects, we never spot a single stuck during our experiments. This is probably because of the handover at the very beginning. The planner reduced the uncertainty using handover and roughly aligned the two objects at the pre-assembly poses.
  
\begin{figure}[!htpb]
  \begin{center}
  \includegraphics[width=.99\linewidth]{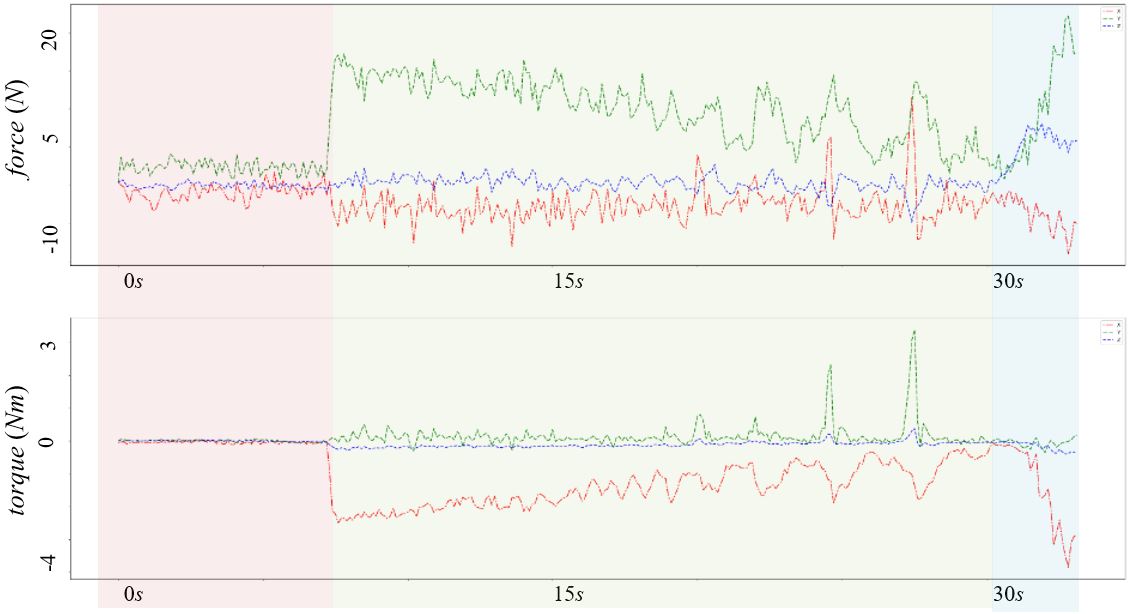}
  \caption{The changes of force (upper diagram) and torque (lower diagram) during the compliant control. The robot is doing linear search in the red time section, spiral search in the green time section, and impedance control int the blue time section.}
  \label{fig:txyz}
  \end{center}
\end{figure}

Fig.\ref{fig:txyz} shows the changes of force and torque during the compliant control. Likewise, the robot is doing a linear search in the red time section, spiral search in the green time section, and impedance control int the blue time section. There are only noises during the linear search. The measured force is less than 4$N$, which is acceptable according to the noise range provided by the F/T300 manual (\rev{0$\sim$1.2$N$ in $x$ and $y$; 0$\sim$0.5$N$ in $z$}). The robot feels a large force in $y$ direction in the green time section and thus continues the spiral search until around 30$seconds$. At this time stop, the force in $y$ falls below our given threshold (7$N$) and the robot switches to impedance control. \rev{The threshold is determined by experience.} Note that the changes in the forces and torques depend on the grasping pose. The $y$ force and $x$ torque in Fig.\ref{fig:txyz} are very significant because the mating hand is holding the object using the pose shown in Fig.\ref{fig:complianceexe}. The hand is moving along the $y$ direction and the object is a bit rotated around the $x$ axis.

\begin{figure}[!htpb]
  \begin{center}
  \includegraphics[width=\linewidth]{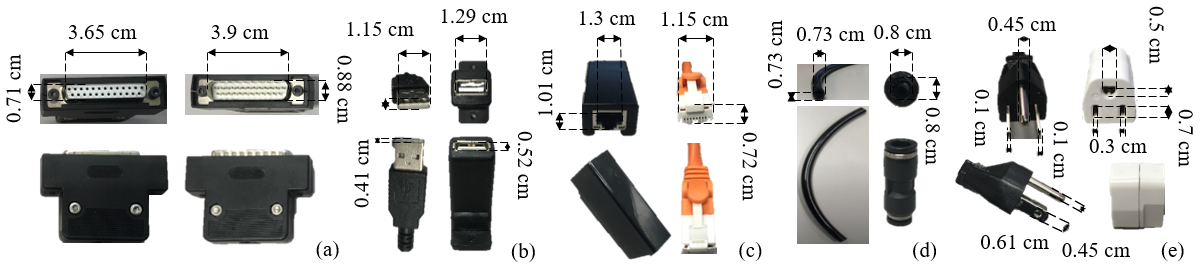}
  \caption{\rev{The additional objects used in the experiments. The left column of each subfigure shows the mating object. The right column shows the assembly object. (a) A D-Sub DB25 connector. (b) A USB connector. (c) A RJ45 internet connector. (d) A one-touch vacuum connector. (e) A three-prong power plug.}}
  \label{fig:othercomponents}
  \end{center}
\end{figure}

\begin{figure}[!htpb]
  \begin{center}
  \includegraphics[width=\linewidth]{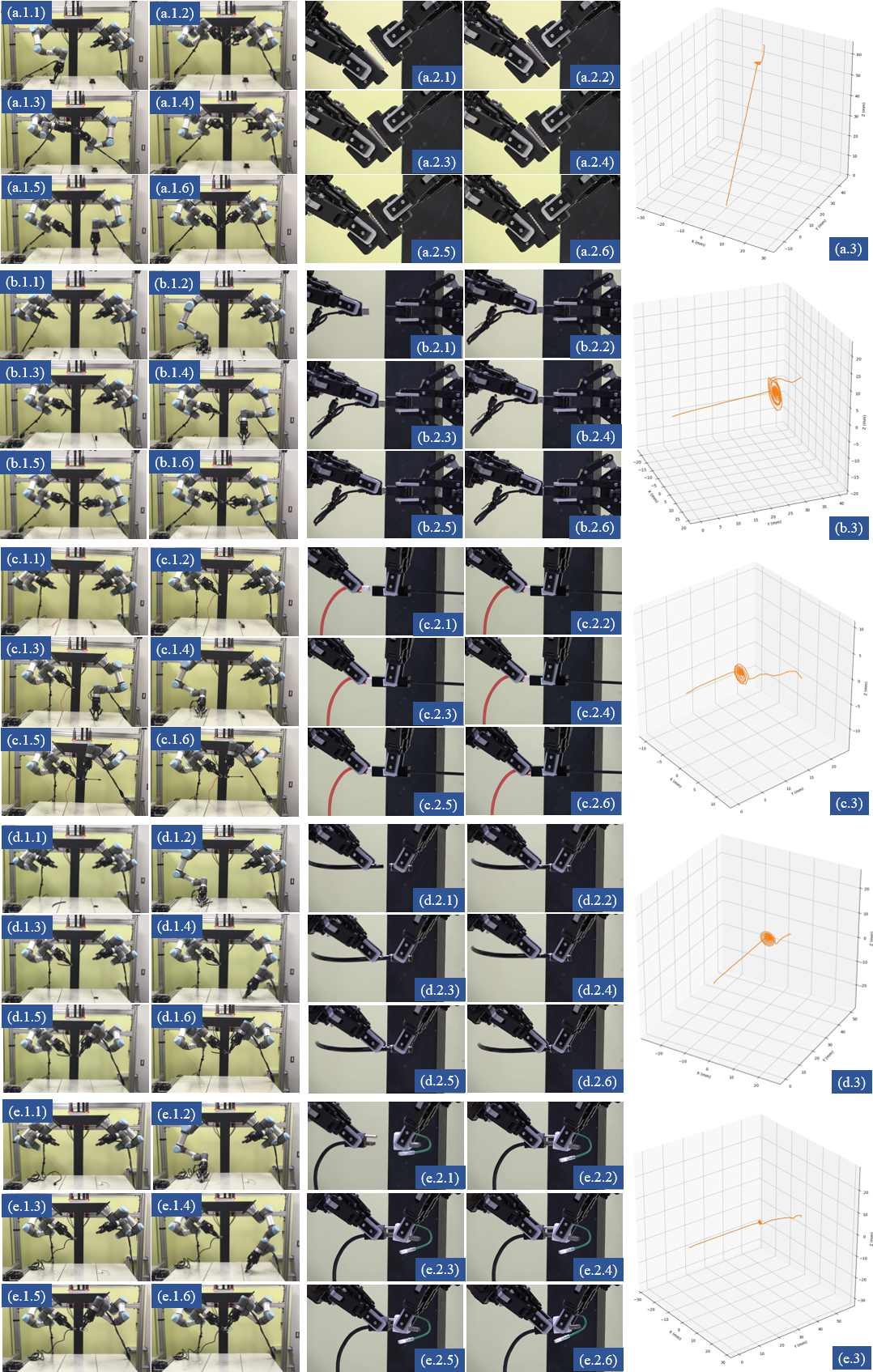}
  \caption{\rev{(Left column) The real execution of picking and assembling the different objects. The (a)-(e) in the figure corresponds to the (a)-(e) objects in Fig.\ref{fig:othercomponents}. (Middle column) A close-up view of the objects in the scene during compliant control. (Right column) The trajectories of the mating gripper during the compliant control.}}
  \label{fig:realworld}
  \end{center}
\end{figure}

\begin{figure}[!htpb]
  \begin{center}
  \includegraphics[width=\linewidth]{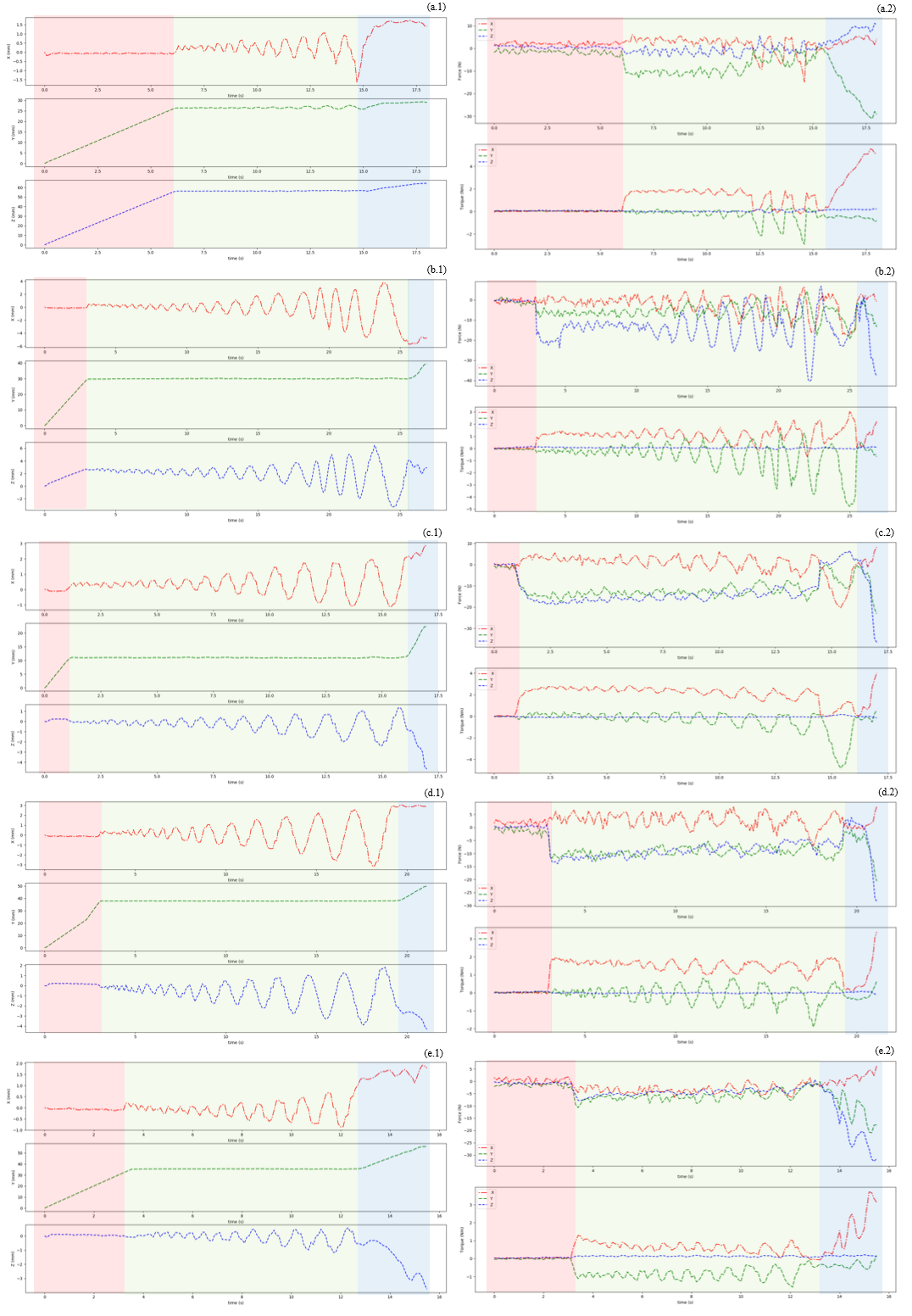}
  \caption{\rev{(Left column) The changes in the $x$, $y$, $z$ positions of the mating gripper during inserting the object in Fig.\ref{fig:othercomponents}. The red time area indicates the linear search. The green time area indicates that the robot is doing the spiral search. The blue time area indicates the impedance control. (Right column) The changes of force (upper diagram) and torque (lower diagram) during the compliant control of the objects mentioned in Fig.\ref{fig:othercomponents}. The robot is doing linear search in the red time section, spiral search in the green time section, and impedance control int the blue time section.}}
  \label{fig:obj_gripper_motion}
  \end{center}
\end{figure}

\rev{Besides the manifold connector, we tested several other tasks to analyze the success rates, the time efficiency, as well as the effort needed to adapt to new workpieces. The tasks include: 1) inserting a multi-pin connector, 2) inserting a USB connector, 3) inserting an internet cable, 4) inserting a vacuum tube, and 5) inserting a power plug. The related objects are shown in Fig.\ref{fig:othercomponents}. Each of the objects has a different hole type and our method could successfully finish all with properly selected parameters. Table \ref{tb:ob_timeconsuming} shows the detailed time costs of each step. The sequences of the planned motion are shown by the snapshots in the left column of Fig.\ref{fig:realworld}. The changes of the objects during the compliant control are shown in the middle and right columns of the same figure. The changes in the $x$, $y$, $z$ positions of the mating gripper are shown in the left column of Fig.\ref{fig:obj_gripper_motion}. The changes of force and torque during the compliant control are shown in the right column of the same figure. The results demonstrate that our method applies to various peg-in-hole tasks. The integration with compliant control can tolerate the large absolute position errors of conventional combined task and motion planners.}

\begin{table}[!htbp]
  \centering
  \caption{\rev{\label{conf0}Time costs of the additional objects}}
  \label{tb:ob_timeconsuming}
  \resizebox{.83\linewidth}{!}{%
  \begin{tabular}{ccc|c|c}\toprule
  
  \multirow{8}{0.17\textwidth}{Inserting the multi-pin connector} & \dunderline{.57pt}{Mating object} &  Pick up &  Regrasp 1 & Regrasp 2 \\
  ~ & ~ & 7$s$     & 25$s$            & 28$s$   \\ \cmidrule{3-5}
  ~ & ~ & Regrasp 3 & Regrasp 4   & Goal pose \\
  ~ & ~ & 21$s$ & 24$s$             & 6$s$ \\ \cmidrule{2-5}
  ~ & \dunderline{.57pt}{Assembly object} & Grasp & Goal pose & - \\ 
  ~& ~ & 9$s$    & 13$s$  & \\  \cmidrule{2-5}
  ~ & \dunderline{.57pt}{Compliant control} & Linear & Spiral & Impedance\\ 
  ~& ~& 6$s$    & 9$s$ & 3$s$\\
  \midrule
  \multirow{6}{0.17\textwidth}{Inserting the USB connector} & \dunderline{.57pt}{Mating object} &  Pick up & Goal pose &\\
  ~ &~ & 12$s$             & 11$s$             &  \\ \cmidrule{2-5}
  ~ & \dunderline{.57pt}{Assembly object} & Grasp & Goal pose & -  \\ 
  ~& ~ & 9$s$    & 9$s$  & \\  \cmidrule{2-5}
  ~ & \dunderline{.57pt}{Compliant control} & Linear & Spiral & Impedance  \\ 
  ~& ~& 4$s$    & 26$s$ & 2$s$ \\
  \midrule
  \multirow{6}{0.17\textwidth}{Inserting the internet cable} & \dunderline{.57pt}{Mating object} &  Pick up & Goal pose &\\
  ~ &~ & 8$s$             & 9$s$             & \\ \cmidrule{2-5}
  ~ & \dunderline{.57pt}{Assembly object} & Grasp & Goal pose & -\\ 
  ~& ~ & 8$s$    & 10$s$  & \\  \cmidrule{2-5}
  ~ & \dunderline{.57pt}{Compliant control} & Linear & Spiral & Impedance \\ 
  ~& ~& 2$s$    & 15$s$ & 3$s$ \\
  \midrule
  \multirow{6}{0.17\textwidth}{Inserting the vacuum tube} & \dunderline{.57pt}{Mating object} &  Pick up & Goal pose &\\
  ~ &~ & 8$s$             & 10$s$             &   \\ \cmidrule{2-5}
  ~ & \dunderline{.57pt}{Assembly object} & Grasp & Goal pose & - \\ 
  ~& ~ & 9$s$    & 13$s$  & \\  \cmidrule{2-5}
  ~ & \dunderline{.57pt}{Compliant control} & Linear & Spiral & Impedance \\ 
  ~& ~& 3$s$    & 25$s$ & 2$s$ \\
  \midrule
  \multirow{6}{0.17\textwidth}{Inserting the power plug} & \dunderline{.57pt}{Mating object} &  Pick up & Goal pose &\\
  ~ &~ & 9$s$             & 10$s$             & \\ \cmidrule{2-5}
  ~ & \dunderline{.57pt}{Assembly object} & Grasp & Goal pose & -  \\ 
  ~& ~ & 8$s$    & 11$s$  & \\  \cmidrule{2-5}
  ~ & \dunderline{.57pt}{Compliant control} & Linear & Spiral & Impedance \\ 
  ~& ~& 4$s$    & 12$s$ & 3$s$ \\
  \midrule
  \end{tabular}  
  }
\end{table}

\rev{To adapt the planner to new objects, we only need to replace the models of the new objects and indicate the final assembly position. The planner will automatically generate the motions to move the object into the specified position and execute the insertion. Note that we predefined a threshold for the spiral search and the parameters for the impedance control from experience. The predefined values render the robot to perform searching and insertion motion mildly and thus avoid damaging the objects. For some insertions that need more force and torque, for example inserting the plug into the socket, one needs to increase the damping coefficient to ensure successful executions.}

\section{Conclusions and Future Work}

In this paper, we proposed a method that integrates combined task and motion planning with compliant peg-in-hole control to successfully \rev{conduct} planned dual-arm assembly motion. The method is used by a dual-arm UR3 robot to assemble a complex connector made of fifteen peg-hole pairs. The results show that the method has high robustness and high fidelity. It is effective for robots with planned motion. The combined task and motion planning are important to move an object from an arbitrary pose to a pre-assembly pose. The compliant strategies then help to avoid offsets from both calibration errors and grasp uncertainty. They together enable a robot to finish up and whole pick-up-and-insertion task.

Our future work includes leveraging machine learning techniques to increase time efficiency, as well as dealing with
soft or changeable objects.

\begin{acknowledgements}
This paper is based on results obtained from a project commissioned by the New Energy and Industrial Technology Development Organization (NEDO).
\end{acknowledgements}

\bibliographystyle{spphys}       
\bibliography{paper.bib}   

\begin{thebibliography}{10}
\providecommand{\url}[1]{{#1}}
\providecommand{\urlprefix}{URL }
\expandafter\ifx\csname urlstyle\endcsname\relax
  \providecommand{\doi}[1]{DOI \discretionary{}{}{}#1}\else
  \providecommand{\doi}{DOI \discretionary{}{}{}\begingroup
  \urlstyle{rm}\Url}\fi

\bibitem{quigley2009ros}
M.~Quigley, K.~Conley, B.~Gerkey, J.~Faust, T.~Foote, J.~Leibs, R.~Wheeler,
  A.Y. Ng, in \emph{ICRA workshop on open source software}, vol.~3 (Kobe,
  Japan, 2009), vol.~3, p.~5

\bibitem{Thierry04}
T.~Simeon, J.P. Laumond, J.~Cortes, A.~Shbani, International Journal of
  Robotics Research \textbf{23}, 729 (2004)

\bibitem{cowley2013perception}
A.~Cowley, B.~Cohen, W.~Marshall, C.J. Taylor, M.~Likhachev, in \emph{2013
  IEEE/RSJ International Conference on Intelligent Robots and Systems} (IEEE,
  2013), pp. 816--823

\bibitem{domae2014fast}
Y.~Domae, H.~Okuda, Y.~Taguchi, K.~Sumi, T.~Hirai, in \emph{2014 IEEE
  International Conference on Robotics and Automation (ICRA)} (IEEE, 2014), pp.
  1997--2004

\bibitem{harada2014validating}
K.~Harada, T.~Tsuji, K.~Nagata, N.~Yamanobe, H.~Onda, Robotics and Autonomous
  Systems \textbf{62}(10), 1463 (2014)

\bibitem{shirai1973guiding}
Y.~Shirai, H.~Inoue, Pattern Recognition \textbf{5}(2), 99 (1973)

\bibitem{yoshimi1994active}
B.H. Yoshimi, P.K. Allen, in \emph{ICRA} (1994), pp. 156--161

\bibitem{huang2013fast}
S.~Huang, et~al., in \emph{IROS} (2013), pp. 286--292

\bibitem{song2014automated}
H.C. Song, Y.L. Kim, J.B. Song, in \emph{2014 IEEE/RSJ International Conference
  on Intelligent Robots and Systems} (IEEE, 2014), pp. 4517--4522

\bibitem{li2017human}
X.~Li, R.~Li, H.~Qiao, C.~Ma, L.~Li, in \emph{2017 IEEE/RSJ International
  Conference on Intelligent Robots and Systems (IROS)} (IEEE, 2017), pp.
  4743--4748

\bibitem{newman2001interpretation}
W.S. Newman, et~al., in \emph{ICRA} (2001), pp. 571--576

\bibitem{chhatpar2001search}
S.R. Chhatpar, M.S. Branicky, in \emph{IROS}, vol.~3 (2001), vol.~3, pp.
  1465--1470

\bibitem{park2017compliance}
H.~Park, J.~Park, D.H. Lee, J.H. Park, M.H. Baeg, J.H. Bae, IEEE Transactions
  on Industrial Electronics \textbf{64}(8), 6299 (2017)

\bibitem{marvel2018multi}
J.A. Marvel, R.~Bostelman, J.~Falco, ACM Computing Surveys (CSUR)
  \textbf{51}(1), 14 (2018)

\bibitem{abdullah2015approach}
M.W. Abdullah, H.~Roth, M.~Weyrich, J.~Wahrburg, IFAC-PapersOnLine
  \textbf{48}(3), 1476 (2015)

\bibitem{whitney1982quasi}
D.E. Whitney, J. Dyn. Sys., Meas., Control. \textbf{104}(1), 65 (1982)

\bibitem{haskiya1998passive}
W.~Haskiya, K.~Maycock, J.~Knight, Proceedings of the Institution of Mechanical
  Engineers, Part B: Journal of Engineering Manufacture \textbf{212}(6), 473
  (1998)

\bibitem{zhao1998vrcc}
F.~Zhao, P.S. Wu, Mechatronics \textbf{8}(6), 657 (1998)

\bibitem{joo1998development}
S.~Joo, F.~Miyazaki, in \emph{Proceedings. 1998 IEEE/RSJ International
  Conference on Intelligent Robots and Systems. Innovations in Theory, Practice
  and Applications (Cat. No. 98CH36190)}, vol.~2 (IEEE, 1998), vol.~2, pp.
  1326--1332

\bibitem{lee2005development}
S.~Lee, IEEE Transactions on Automation Science and Engineering \textbf{2}(2),
  193 (2005)

\bibitem{dreaming}
T.~Nishimura, Y.~Suzuki, T.~Tsuji, T.~Watanabe, in \emph{ICHR} (2017), pp.
  67--74

\bibitem{balletti2012towards}
L.~Balletti, A.~Rocchi, F.~Belo, M.~Catalano, M.~Garabini, G.~Grioli,
  A.~Bicchi, in \emph{2012 12th IEEE-RAS International Conference on Humanoid
  Robots (Humanoids 2012)} (IEEE, 2012), pp. 402--408

\bibitem{song2016guidance}
H.C. Song, Y.L. Kim, J.B. Song, Advanced Robotics \textbf{30}(8), 552 (2016)

\bibitem{hogan1987stable}
N.~Hogan, in \emph{Proceedings. 1987 IEEE International Conference on Robotics
  and Automation}, vol.~4 (IEEE, 1987), vol.~4, pp. 1047--1054

\bibitem{liu2019}
S.~{Liu}, D.~{Xing}, Y.~{Li}, J.~{Zhang}, D.~{Xu}, IEEE/ASME Transactions on
  Mechatronics \textbf{24}(5), 1974 (2019)

\bibitem{nammoto2013model}
T.~Nammoto, K.~Kosuge, K.~Hashimoto, in \emph{2013 IEEE International
  Conference on Automation Science and Engineering (CASE)} (IEEE, 2013), pp.
  948--953

\bibitem{wally2019flexible}
B.~Wally, J.~Vysko{\v{c}}il, P.~Nov{\'a}k, C.~Huemer,
  R.~{\v{S}}indel{\'a}{\v{r}}, P.~Kadera, A.~Mazak, M.~Wimmer, IEEE Robotics
  and Automation Letters \textbf{4}(4), 4062 (2019)

\bibitem{zhang2017multirobot}
S.~Zhang, Y.~Jiang, G.~Sharon, P.~Stone, in \emph{Proceedings of the 16th
  Conference on Autonomous Agents and MultiAgent Systems} (International
  Foundation for Autonomous Agents and Multiagent Systems, 2017), pp. 501--510

\bibitem{ratliff2009chomp}
N.~Ratliff, M.~Zucker, J.A. Bagnell, S.~Srinivasa,   (2009)

\bibitem{schulman2014motion}
J.~Schulman, Y.~Duan, J.~Ho, A.~Lee, I.~Awwal, H.~Bradlow, J.~Pan, S.~Patil,
  K.~Goldberg, P.~Abbeel, The International Journal of Robotics Research
  \textbf{33}(9), 1251 (2014)

\bibitem{Lavalle00}
S.M. Lavalle, J.J. Kuffner, in \emph{Proceedings of International Workshop on
  the Algorithmic Foundations of Robotics} (2000), pp. 293--308

\bibitem{wolfe2010combined}
J.~Wolfe, B.~Marthi, S.~Russell, in \emph{Twentieth International Conference on
  Automated Planning and Scheduling} (2010)

\bibitem{Kaelbling13}
L.P. Kaelbling, T.~Lozano-Perez, International Journal of Robotics Research
  (2013)

\bibitem{garrett2015ffrob}
C.R. Garrett, T.~Lozano-P{\'e}rez, L.P. Kaelbling, in \emph{Algorithmic
  Foundations of Robotics XI} (Springer, 2015), pp. 179--195

\bibitem{woosley2018integrated}
B.~Woosley, P.~Dasgupta, Robotica \textbf{36}(3), 353 (2018)

\bibitem{Wan2015a}
W.~Wan, M.T. Mason, R.~Fukui, Y.~Kuniyoshi, in \emph{Proceedings of
  International Conference on Robotics and Automation} (2015), pp. 4326--4333

\bibitem{wan2019tii}
W.~{Wan}, K.~{Harada}, F.~{Kanehiro}, IEEE Transactions on Industrial
  Informatics pp. 1--1 (2019)

\bibitem{Wan2016tdar}
W.~Wan, K.~Harada, Advanced Robotics pp. 1111--1125 (2016)

\bibitem{moriyama2019dual}
R.~Moriyama, W.~Wan, K.~Harada, arXiv preprint arXiv:1903.00646  (2019)

\bibitem{siciliano2012advanced}
B.~Siciliano, \emph{Advanced bimanual manipulation: Results from the dexmart
  project}, vol.~80 (Springer Science \& Business Media, 2012)

\bibitem{kruger2011dual}
J.~Kr{\"u}ger, G.~Schreck, D.~Surdilovic, CIRP annals \textbf{60}(1), 5 (2011)

\bibitem{cohen2015planning}
B.~Cohen, M.~Phillips, M.~Likhachev, in \emph{Eighth Annual Symposium on
  Combinatorial Search} (2015)

\bibitem{kurosu2017simultaneous}
J.~Kurosu, A.~Yorozu, M.~Takahashi, Applied Sciences \textbf{7}(12), 1210
  (2017)

\bibitem{ramirez2017human}
I.G. Ramirez-Alpizar, K.~Harada, E.~Yoshida, Robomech Journal \textbf{4}(1), 20
  (2017)

\bibitem{hwang2007motion}
M.J. Hwang, D.Y. Lee, S.Y. Chung, in \emph{2007 IEEE International Conference
  on Systems, Man and Cybernetics} (IEEE, 2007), pp. 240--245

\bibitem{stavridis2018bimanual}
S.~Stavridis, Z.~Doulgeri, in \emph{2018 IEEE/RSJ International Conference on
  Intelligent Robots and Systems (IROS)} (IEEE, 2018), pp. 7131--7136

\bibitem{rodrigues1840lois}
R.~Askey, Mathematics and Social Utopias in France: Olinde Rodrigues and His
  Times \textbf{28}, 105 (2005)

\end{thebibliography}

\end{document}